\documentclass[11pt]{article}
\usepackage[margin=1in]{geometry}
\usepackage{times}
\usepackage{authblk}
\usepackage{booktabs}\usepackage{xspace}\usepackage{subcaption}\usepackage{enumitem}\usepackage{graphicx}
\usepackage[numbers,sort&compress]{natbib}
\usepackage[capitalize,noabbrev]{cleveref}
\usepackage{microtype}
\usepackage{xcolor}
\graphicspath{{figures/}}

\newcommand{\system}{\textsc{Rigel}\xspace}
\newcommand{\etal}{\textit{et al.}\xspace}
\newcommand{\para}[1]{\smallskip\noindent\textbf{#1}\xspace}
\newcommand{\TF}{\,TFLOP/s\xspace}

\title{\bfseries \system: Reverse-Engineering the Metal 4.1 Tensor Compute Path\\ on the Apple M4 Max GPU}
\author[ ]{Ramchand Kumaresan}
\affil[ ]{Murai Labs \\ \texttt{ramchand@murailabs.com}}
\date{June 2026}

\begin{document}
\maketitle

\begin{abstract}
\noindent
Apple's Metal~4.1 exposes a tensor compute path: the Metal Performance Primitives (MPP)
\texttt{matmul2d} operation over \texttt{cooperative\_tensor} fragments, whose \emph{interface} is
documented but whose \emph{hardware behavior is deliberately hidden}. The specification states which
data-type rows are supported, never whether they are hardware-accelerated, where the operation
physically executes, what its accumulator width is, or how it partitions matrix fragments across
threads. We present \system, an empirical characterization of this path on a single Apple M4 Max
(a pre--neural-accelerator generation). Using a checksum-gated, provenance-tracked microbenchmark
harness, \system recovers eleven facts the v4.1 specification hides or contradicts. The headline
finding: the Metal~4.1 fp8 (E4M3) \texttt{matmul2d} is \emph{emulated}, not accelerated: it sustains
$0.94\times$ the throughput of fp16 despite reading half the operand bytes, so on M4 it is a
memory-footprint feature, not a performance feature. We further show, via a three-signal triangulation
(throughput ceiling, comparison against \texttt{simdgroup\_matrix}, and per-rail power attribution),
that \texttt{matmul2d} executes entirely on the GPU shader cores with no dedicated matrix datapath and
no evidence of Apple Neural Engine routing; that it accumulates in $\geq$fp32; and we reconstruct the opaque
$8{\times}8$ \texttt{cooperative\_tensor} fragment layout Apple documents nowhere. Acting on the
characterization, a hand-fused GEMM\,+\,bias\,+\,GELU kernel beats the decomposed path by
$+6.5$--$12.9\%$ in the cache-resident regime. All findings are reproducible from committed MIT-licensed
code and per-cell CSVs.
\end{abstract}

\section{Introduction}\label{sec:intro}

Quantized and attention-heavy machine-learning workloads increasingly run on consumer Apple Silicon
GPUs. Metal~4.1 (shipping in Xcode~27 / macOS~27.0) adds a tensor compute path (the Metal
Performance Primitives (MPP) \texttt{matmul2d} operation, which contracts two \texttt{tensor} operands
through device-distributed \texttt{cooperative\_tensor} fragments) and a low-precision frontier of
fp8, fp4, and block-scaled MXFP4 formats. For anyone deploying an LLM or vision model on a Mac, three
questions decide everything: \emph{where} does a tensor op execute, \emph{how fast} is each
low-precision format, and \emph{what} numerical guarantees does it provide. The Metal Shading Language
Specification~\cite{metalspec} answers none of them.

\para{The problem:} The specification documents the \emph{interface} but hides the \emph{hardware
behavior}. Its feature tables state that a data-type row is \emph{supported}; they never state whether
it is \emph{accelerated}. It declares the \texttt{cooperative\_tensor} layout ``opaque'' and ``device
specific'' (\S2.22.3.1). It is silent on the dispatch target of \texttt{matmul2d}, on accumulator
width, and on the microarchitectural alignment constraints a kernel must satisfy. A developer cannot
tell, from the specification alone, whether fp8 will be twice as fast as fp16 or no faster at all.

\para{The gap:} No public source characterizes the Metal~4.1 tensor path on a specific Apple GPU.
Workload-level studies benchmark Apple Silicon end to end~\cite{fengapple,hubnerapple} but do not
open the tensor primitive itself. The community is split on a load-bearing question: whether Apple's
only ``tensor core'' is the legacy \texttt{simdgroup\_matrix} instruction (an ALU-utilization
optimization) or a dedicated matrix unit analogous to NVIDIA tensor cores~\cite{markidis,suntensorcores}.
The M4 Max sits \emph{before} the GPU ``Neural Accelerators'' Apple reports for the A19/M5 family.

\para{The insight:} These hidden facts are \emph{recoverable empirically}, in the microbenchmarking
tradition that dissected NVIDIA GPUs~\cite{jiavolta,jiaturing}, and on a pre--neural-accelerator chip
the strong prior is that the fp8/fp4 rows are functionally supported but \emph{emulated}. We adopt a
discovery bar: a result counts only if it is \textsc{hidden} or \textsc{contradicted} relative to the
specification; facts a careful reader could derive from the specification are controls, not contributions.

\para{Contributions:}
\begin{enumerate}[leftmargin=*,topsep=2pt,itemsep=1pt]
\item A checksum-gated, provenance-tracked harness (\cref{sec:method}) that turns each claim into a
reproducible cell with a float64 reference and a pre-registered cheap-baseline falsification gate.
\item The \textsc{headline}: fp8 (E4M3) \texttt{matmul2d} is \emph{emulated} on M4 Max ($0.94\times$
fp16 throughput), making it a footprint feature, not a perf feature (\cref{sec:h4}).
\item A three-signal dispatch-target result (\cref{sec:h1}): \texttt{matmul2d} runs on the GPU shader
cores via the legacy \texttt{simdgroup\_matrix} path, with no dedicated matrix unit and no evidence of
ANE routing.
\item Reconstruction of the opaque $8{\times}8$ \texttt{cooperative\_tensor} fragment layout
(\cref{sec:h3}), plus the accumulator width ($\geq$fp32) and the fp8/fp4 numeric semantics.
\item A \textsc{contradicted}-class version gate: Metal~4.1 requires Xcode~27 \emph{and} macOS~27.0 (not
the specification's ``Xcode~26.1+''), and a 4.1 binary refuses to load on macOS~26.5 (\cref{sec:bg}).
\item The eleven hidden/contradicted findings as a single legible catalog (\cref{tab:findings}), the
natural home for low-visibility constraints (the 128-byte alignment, the SFINAE footgun, the version gate).
\item An optimization study (\cref{sec:opt}): a hand-fused GEMM+bias+GELU kernel beats the decomposed
path by $+6.5$--$12.9\%$ in the cache-resident regime (the win decays as $O(n^3)$ compute dominates),
while a fused-attention study shows the win does not extend to operations that forgo the matrix units.
\end{enumerate}

\begin{table}[t]\centering\small
\renewcommand{\arraystretch}{1.35}
\setlength{\tabcolsep}{8pt}
\caption{The eleven findings that meet our discovery bar: each is either \textsc{hidden} by the
Metal~4.1 specification or directly \textsc{contradicts} it, and none can be derived from the document
alone. Eight are hidden (H) and three are contradicted (C). Controls, such as the OCP-conformant fp8 and
fp4 grids, are excluded. The ``Spec.\ basis'' column names what the specification says, or fails to say,
about each point.}
\label{tab:findings}
\begin{tabular}{@{}p{0.575\linewidth} c l l@{}}
\toprule
\textbf{Finding} & \textbf{Class} & \textbf{Spec.\ basis} & \textbf{Evidence} \\
\midrule
fp8 (E4M3) \texttt{matmul2d} is emulated, not accelerated ($0.94\times$ fp16) & H & feature table & \cref{fig:h4} \\
\texttt{matmul2d} runs on the GPU shader cores; no dedicated matrix unit & H & silent & \cref{fig:h1signals} \\
Accumulator width is at least fp32 & H & silent & \cref{fig:accum} \\
\texttt{cooperative\_tensor} layout is an $8{\times}8$ fragment, unified across $A$/$B$/$C$ & H & \S2.22.3.1 & \cref{fig:frag} \\
Metal 4.1 needs Xcode 27 \emph{and} macOS 27.0; a 4.1 binary will not load on 26.5 & C & ``Xcode 26.1+'' & \cref{sec:bg} \\
A host-bound \texttt{tensor\_inline} inner stride must be 128-byte aligned & H & SFINAE note & \cref{sec:semantics} \\
\texttt{tensor\_inline} cannot tile a GEMM (\texttt{slice()} does not offset reads) & H & silent & \cref{sec:opt} \\
A device \texttt{tensor} is host-bound only (no pointer constructor) & H & silent & \cref{sec:bg} \\
Tensors are gated on \texttt{\_\_HAVE\_TENSOR\_\_} and deployment target $\geq$26.2 & H & silent & \cref{sec:bg} \\
The 4.1 low-precision float formats are absent under the 4.0 toolchain & C & feature table & \cref{sec:bg} \\
\texttt{int2b} and block-scaled \texttt{tensor\_blockwise} (MXFP4) are absent & C & feature table & \cref{sec:bg} \\
\bottomrule
\end{tabular}
\end{table}

\section{Background}\label{sec:bg}

This section gives the reader the three things needed to follow the results: what the Metal~4.1 tensor
path actually is, the version gate that controls access to it, and the device on which we measure.

\para{The Metal 4.1 tensor path:} The primitive at the centre of this paper is the Metal Performance
Primitives operation \texttt{matmul2d}, which multiplies two matrices into a result, $C = A\cdot B$. Its
operands are \texttt{tensor} objects, and Metal offers two flavours that matter here. A
\texttt{tensor\_inline} is a lightweight, non-owning view over an ordinary \texttt{device} buffer,
created inside the shader and describing only a contiguous region; a host-bound \texttt{tensor\_handle}
is instead constructed on the CPU and can carry explicit strides. The destination may additionally be
declared a \texttt{cooperative\_tensor}, a register fragment that the threads of an execution scope
hold collectively, never writing it to device memory, and whose internal layout the specification
deliberately leaves opaque. Each call is shaped by a descriptor, \texttt{matmul2d\_descriptor(m,n,k)},
that fixes the output tile a single threadgroup produces. Passing \texttt{dynamic\_extent} as the
contraction length $k$ tells the operation to loop over that dimension internally, rather than expecting
the caller to drive it. The execution scope, written \texttt{execution\_simdgroups<N>}, then binds $N$
SIMD-groups of 32 threads each to compute one tile cooperatively. These few types, and the hardware behaviours the specification omits for
them, are the entire subject of the paper.

\para{The version gate:} Before any tensor feature can be measured, a binary that uses it must compile
\emph{and} run, and here the specification is already inaccurate. It states that Metal~4.1 is available
from Xcode~26.1 onward, but we found access to be gated on both the toolchain and the operating system.
Xcode~26.5 and the 26.6 release candidate still emit Metal~4.0; only Xcode~27 produces the
language-version marker \texttt{\_\_METAL\_VERSION\_\_\,410}. A library compiled at version 4.1 under
Xcode~27 then refuses to load on macOS~26.5, failing at runtime with ``language version 4.1 is not
supported on this OS.'' Exercising the fp8 and fp4 paths therefore requires Xcode~27 and macOS~27.0
together, not Xcode~26.1 as documented. This is the first of the contradicted-class findings catalogued
in \cref{tab:findings}.

\para{Device under test:} All measurements are taken on a single Apple M4 Max (40 GPU cores, 64\,GB of
unified memory, in a Mac~Studio) running macOS~27.0, build 26A5353q. The M4 generation predates the GPU
``Neural Accelerators'' that Apple introduced with the A19 and M5 parts, so the expectation going in is a
chip that supports the low-precision formats functionally but does not accelerate them; the results bear
this out. Every number is reported ``on M4 Max'', we make no claim about other Apple GPUs, and we treat
cross-generation behaviour as future work.

\section{Methodology}\label{sec:method}

A characterization is only as trustworthy as its defences against fooling itself, so before presenting
any result we describe how \system earns confidence in a measurement. The harness is split across two
languages on purpose. Thin Objective-C++ hosts do nothing but allocate buffers, dispatch a Metal kernel,
and time it, while a dependency-light Python layer owns everything that has to be checked: the schema of
a measurement cell, the per-data-type error tolerances, the correctness gate, and the provenance record.
A number reaches this paper only after passing the four disciplines below.

\para{The checksum gate:} The first question about any GPU kernel is whether it computed the right
answer, and low precision makes that question subtle. An fp8 result differs from a double-precision
reference for two reasons (the \emph{inputs} were quantized, and the \emph{accumulation} lost
precision), but only the second tells us anything about the hardware. We therefore compute the reference
in float64 from the exact same quantized bytes the GPU consumed, so the input-quantization error cancels
and the residual bounds the kernel's accumulation error alone. Any cell whose error exceeds its
per-data-type tolerance is marked \textsc{garbage} and dropped from every average rather than quietly
blended in. As an independent cross-check, a separate Python re-verifier re-derives each pass or fail
verdict directly from the committed CSV; run against real harness output, it agreed on all fifty cells
with zero mismatches.

\para{Provenance:} For a result to be citable months later it has to be reconstructible, so before a run
begins the harness records a configuration hash, the code commit, a hash of the input-generation recipe,
the random seed, and a full capture of the environment (device, GPU core count, OS build, and toolchain
versions). A run that is missing any of these is not used. Every figure in the paper is then regenerated
from its committed CSV by a single \texttt{make reproduce} target that needs no GPU, so a reader can
redraw our plots from our data.

\para{The cheap-baseline gate:} The central risk in any ``the tensor path is fast'' story is that the
speed comes from something mundane (a better loop, or simply the ordinary ALU) rather than from
special hardware. We guard against this by refusing to make a performance claim until the simplest
non-mechanism explanation has actually been run and has failed to account for the effect. Four baselines
stand in for those mundane explanations: (A)~the scalar-ALU roofline, (B)~a naive threadgroup-tiled GEMM
that uses no tensor primitive, (C)~the legacy \texttt{simdgroup\_matrix} instruction, and (D)~the stock
fp16 MPP path. The thresholds are fixed in advance: a kernel must beat the strongest baseline by at least
$1.10\times$ to count as a win, and a low-precision format must reach $1.5\times$ its matched fp16
throughput to count as \emph{accelerated} rather than merely \emph{emulated}.

\para{Reporting:} Each throughput cell is measured over at least thirty repetitions (here $R{=}300$)
after discarding warm-up runs, and we report the median together with a bootstrapped $95\%$ confidence
interval (seed-fixed, $10^4$ resamples); a single-threadgroup ``does it run'' timing is never reported as
throughput. The confidence interval is not boilerplate. As the next paragraph shows, the M4 GPU clock is
bimodal at small problem sizes, and there the width of the interval is itself the finding.

\begin{figure}[t]\centering
\includegraphics[width=\linewidth]{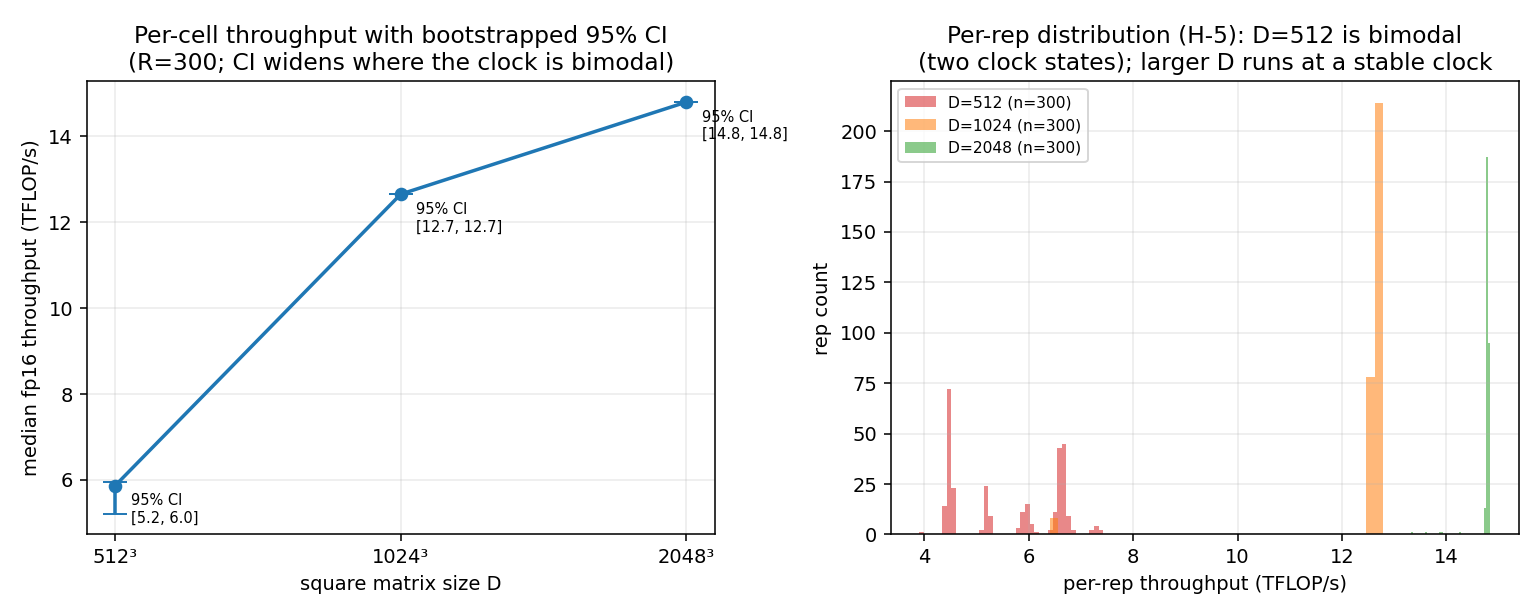}
\caption{\textbf{Clock-state bimodality.} (a) Median fp16 throughput with bootstrapped $95\%$ CI:
the interval is wide at $512^3$ and razor-thin once the clock is stable. (b) Per-rep distributions
($R{=}300$): $512^3$ is bimodal (two clock states), larger sizes are unimodal.}
\label{fig:h5}\end{figure}

That bimodality is itself worth characterizing. Per-rep throughput splits at cache-resident sizes
(\cref{fig:h5}). At $512^3$ the $R{=}300$ samples
split into two near-equal clusters at $\sim$4.65 and $\sim$6.5\TF (145 vs.\ 155 reps), giving a wide
bootstrapped CI of $[5.21, 5.96]$\TF; $1024^3$ runs stably at $\sim$12.7\TF with a rare low-clock tail
($8/300$ reps at $\sim$6.4\TF); $2048^3$ holds a stable max clock ($14.80$\TF, CI $[14.79, 14.80]$). The
mechanism is clock-state: a short kernel finishes around the GPU's clock ramp and lands in either a low
or high P-state, while a long kernel sustains the high state. This clock-state cliff is why every throughput claim in the paper carries a bootstrapped CI rather than
a bare median: a point estimate at $512^3$ would silently report either $4.65$ or $6.5$\TF depending on luck.

\section{Where does \texttt{matmul2d} execute?}\label{sec:h1}

The dispatch target is the load-bearing question: if \texttt{matmul2d} lowers onto the existing GPU
ALU/\texttt{simdgroup\_matrix} pipeline, the fp8 rows cannot be accelerated without dedicated silicon.
We triangulate with three independent signals, the bar the specification's own discovery problem demands.

\begin{figure}[t]\centering
\begin{subfigure}{0.49\linewidth}\centering
\includegraphics[width=\linewidth]{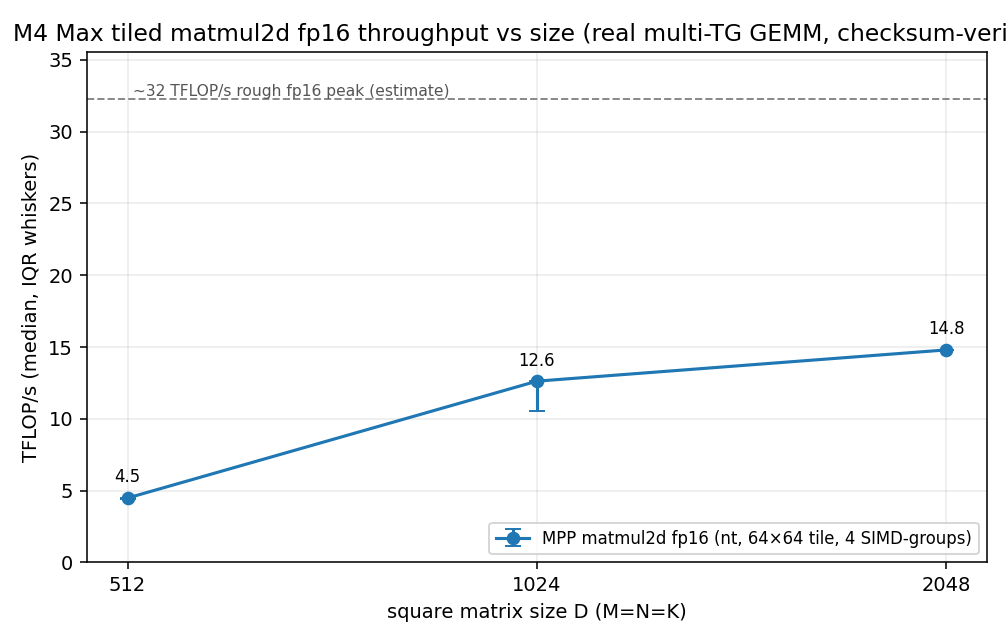}
\caption{fp16 throughput (Signal 1).}\label{fig:throughput}\end{subfigure}\hfill
\begin{subfigure}{0.49\linewidth}\centering
\includegraphics[width=\linewidth]{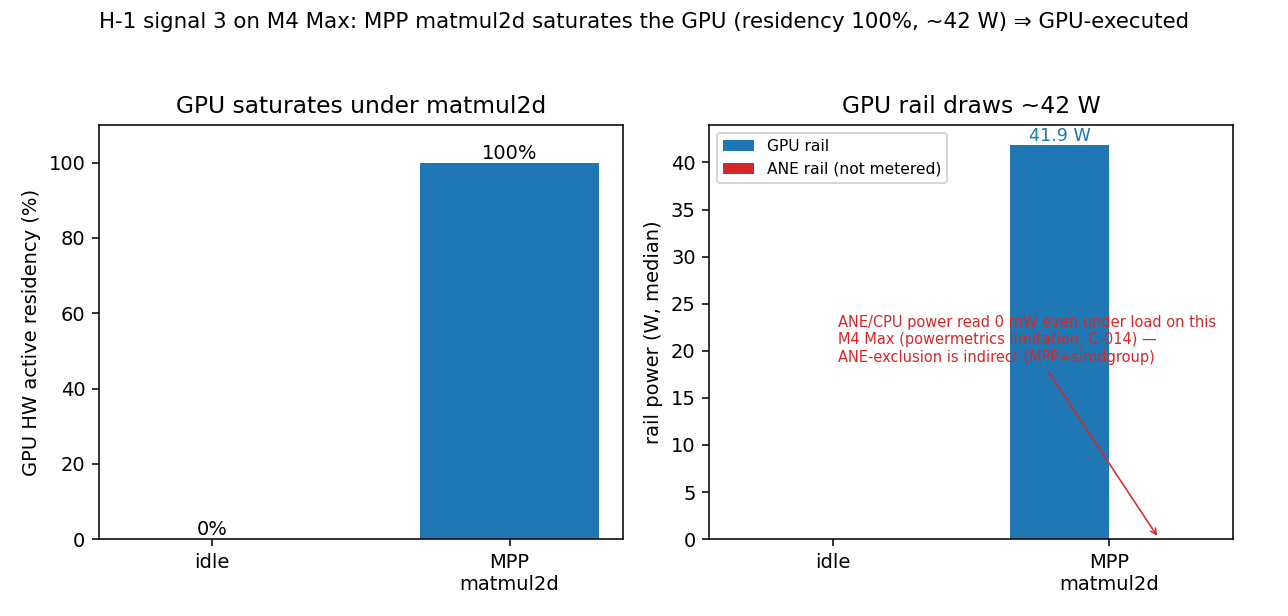}
\caption{GPU power + residency (Signal 3).}\label{fig:power}\end{subfigure}
\caption{\textbf{Dispatch-target signals.} (a) Tiled fp16 \texttt{matmul2d} peaks at $14.8$\TF,
$\approx$46\% of the scalar-ALU roofline; nothing beats the ALU ceiling. Plotted points are medians; the
$512^3$ value is clock-state-dependent (see \cref{fig:h5} for the bimodality). (b) Under a sustained
\texttt{matmul2d} load the GPU rail draws $\sim$42\,W at $100\%$ hardware-active residency.}
\label{fig:h1signals}\end{figure}

\para{Signal 1, throughput ceiling:} A checksum-verified tiled fp16 \texttt{matmul2d} sustains
$14.8$\TF at $2048^3$ (\cref{fig:throughput}), $\approx46\%$ of the $\sim$32\TF scalar-ALU fp16
roofline, never above it, so there is no hidden datapath beating the ALU ceiling.

\begin{figure}[t]\centering
\includegraphics[width=0.74\linewidth]{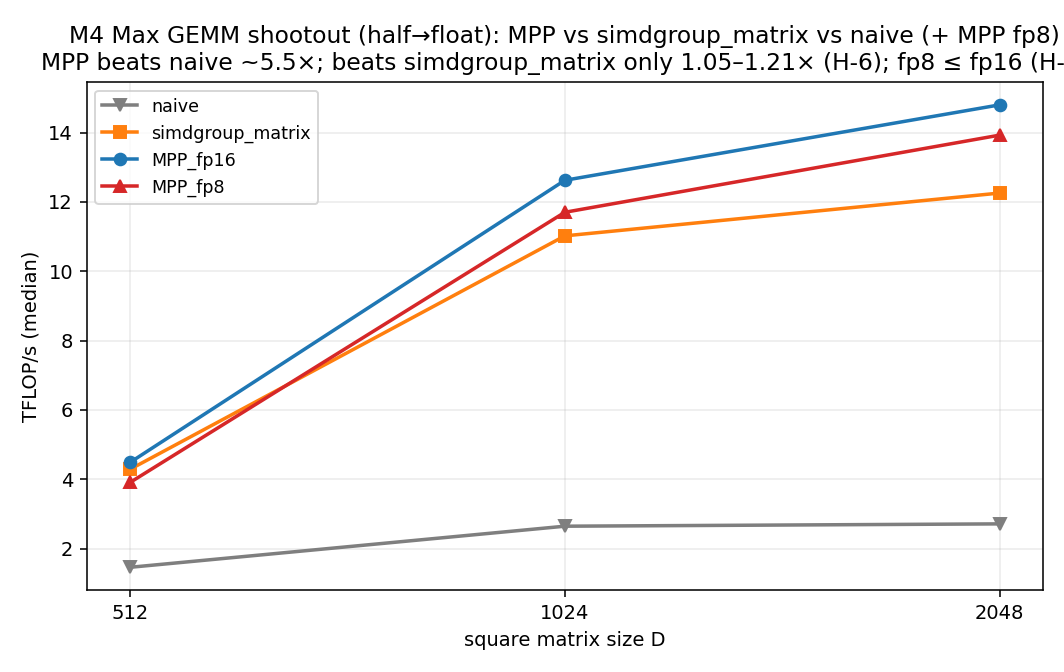}
\caption{\textbf{Four-way GEMM shootout.} Against the deprecated
\texttt{simdgroup\_matrix} path, MPP wins by only $1.05$ to $1.21\times$; against a naive tiled GEMM, by
$2.9$ to $5.5\times$.}
\label{fig:shootout}\end{figure}

\para{Signal 2, comparison:} A four-way GEMM shootout (\cref{fig:shootout}) shows MPP beats a naive
threadgroup-tiled GEMM by $2.9$--$5.5\times$ (the primitive matters), but beats
\texttt{simdgroup\_matrix} by only $1.05$--$1.21\times$. Such proximity is consistent with MPP lowering
onto the \texttt{simdgroup\_matrix} and FP32-ALU path rather than onto a separate unit.

\para{Signal 3, power attribution:} Under a sustained \texttt{matmul2d} load, the GPU rail draws
$\sim$42\,W while the GPU hardware-active-residency hits $100\%$ (\cref{fig:power}); read via a
root-free IOReport ``Energy Model'' reader, the GPU energy rail rises $0.27{\to}47.6$\,W. The
\emph{caveat}: on this macOS~27.0 beta neither \texttt{powermetrics} nor IOReport meters CPU/ANE power
(a CPU at $84\%$ residency still reads $0$\,mW), and no CoreML workload would execute on the ANE
(vision-tower ANE compilation fails), so the ANE-exclusion half of the argument is indirect (it rests
on Signal~2: an ANE offload would \emph{add} throughput beyond what the GPU-only path delivers, and it
does not).

Taken together, these three signals settle the question. On M4 Max, \texttt{matmul2d} executes on the GPU shader cores, on the same
\texttt{simdgroup\_matrix} and FP32-ALU path used by any other shader, with no dedicated matrix datapath. This converts the
contested community claim into a measured fact \emph{on this chip} and is the mechanism behind the fp8
result below.

\begin{figure}[t]\centering
\includegraphics[width=0.74\linewidth]{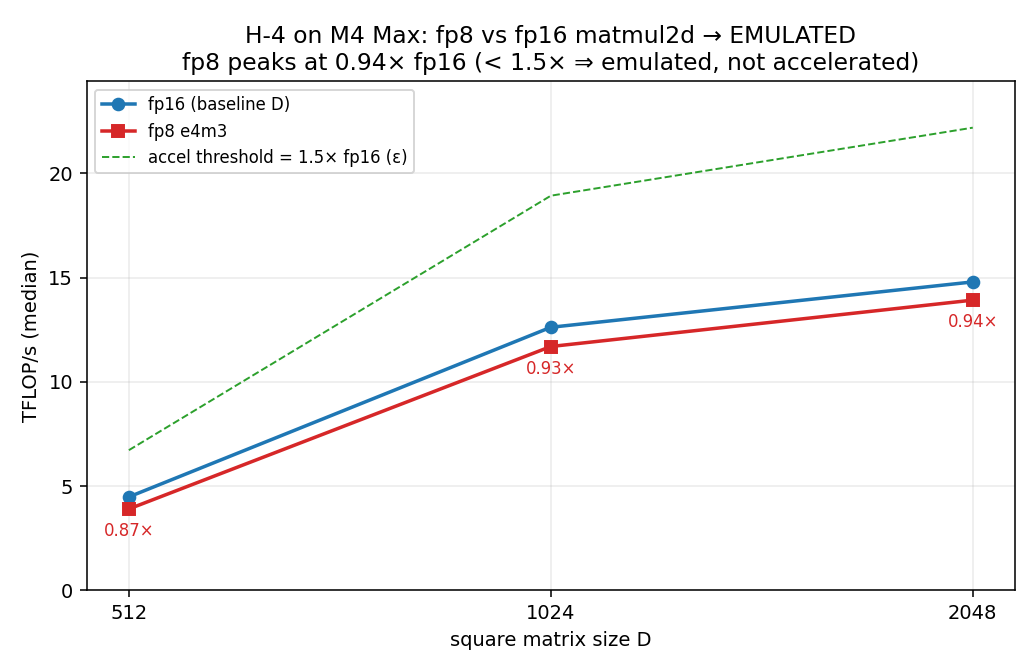}
\caption{\textbf{The headline: fp8 is emulated.} fp8 (E4M3) \texttt{matmul2d} tracks \emph{below} fp16 at every
size (peaks at $0.94\times$), far under the $1.5\times$ acceleration threshold: it is emulated.}
\label{fig:h4}\end{figure}

\section{Is fp8 accelerated or emulated?}\label{sec:h4}

We build identical tiled GEMMs differing only in operand dtype (fp16 vs.\ E4M3), same layout, tile,
and execution scope, both checksum bit-exact. fp8 sustains $0.871$/$0.927$/$0.941\times$ fp16 at
$512$/$1024$/$2048$ (\cref{fig:h4}), peaking at $0.94\times$, nowhere near the pre-registered
$1.5\times$ acceleration bar. The argument is decisive: fp8 operands are one byte versus fp16's two, so
a memory-bound kernel would be \emph{faster} in fp8; fp8 is \emph{slower}, so the matmul is
compute-bound and fp8 carries pure unpack overhead with no low-precision datapath. This is exactly the
prediction for pre--neural-accelerator silicon, and contrasts sharply with Hopper, where fp8 tensor
cores roughly double fp16 throughput~\cite{flashattention3}. \textbf{On M4 Max, the Metal~4.1 fp8
\texttt{matmul2d} is a memory-footprint and dynamic-range feature, not a throughput feature}; the
speedup waits for M5. Per the specification's own framing, a falsified ``faster'' claim is a publishable
characterization finding, and it is the practical result that most changes how a quantized model should
be deployed on M4.

\section{Numeric semantics and hidden constraints}\label{sec:semantics}

Beyond throughput, a developer needs to know how the tensor path behaves numerically: how wide its
accumulator is, what its low-precision formats can represent, and what undocumented constraints a kernel
must satisfy to compile at all. This section settles each in turn.

\begin{figure}[t]\centering
\begin{subfigure}{0.49\linewidth}\centering
\includegraphics[width=\linewidth]{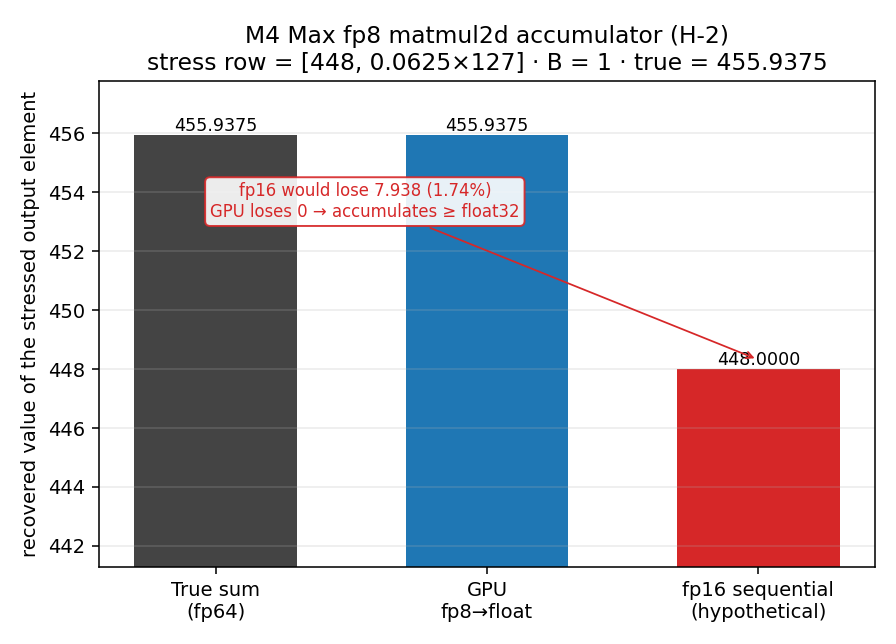}
\caption{Accumulator $\geq$fp32.}\label{fig:accum}\end{subfigure}\hfill
\begin{subfigure}{0.49\linewidth}\centering
\includegraphics[width=\linewidth]{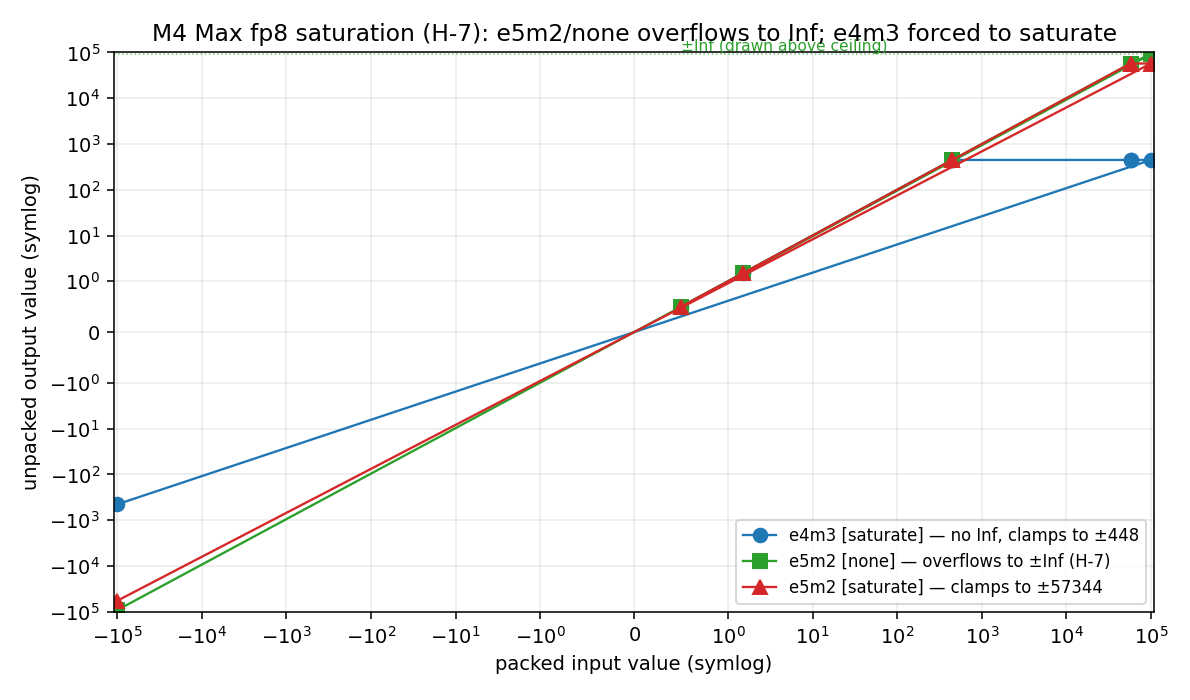}
\caption{E4M3 saturation.}\label{fig:sat}\end{subfigure}
\caption{\textbf{Numeric semantics.} (a) The fp8 \texttt{matmul2d} recovers a sum a sequential fp16
accumulator collapses, so it accumulates in $\geq$fp32. (b) E4M3 clamps to $\pm448$ (no infinity),
even saturating an $+\infty$ input.}
\label{fig:numeric}\end{figure}

The width of the internal accumulator is invisible in an ordinary GEMM, so we expose it with a
deliberately hostile cell. The contraction has depth $K{=}128$ and consists of one large term, $448$,
followed by $127$ small terms of $0.0625$, whose exact sum is $455.9375$. That sum cannot be represented
as a final fp16 value, and a kernel that accumulated in fp16 would lose every small term: the spacing
between representable fp16 values near $448$ is $0.25$, larger than the terms being added, so the running
total would stay pinned at $448$, a $1.74\%$ error. The fp8 \texttt{matmul2d} instead returns $455.9375$
bit-exactly (\cref{fig:accum}), which is only possible if it accumulates in at least fp32. This matches
the at-least-fp32 convention documented for NVIDIA tensor cores~\cite{suntensorcores}, but Apple's
specification never states it for Metal.

The low-precision formats themselves are well behaved. On M4 the fp8 (E4M3, E5M2) and fp4 (E2M1)
representable sets match the OCP specifications~\cite{ocpfp8,ocpmx} exactly, with no divergence across all
$256$ bit patterns. The one behaviour worth flagging is saturation. E5M2 is the only float format whose
default saturation mode is \texttt{none}, so on overflow it yields $\pm\infty$; E4M3, by contrast, has no
infinity encoding and is forced to clamp to $\pm448$, saturating even an $+\infty$ input to $448$
(\cref{fig:sat}).

A more practical surprise is an undocumented alignment rule. A host-bound \texttt{tensor\_inline} requires
its inner stride to be aligned to $128$ \emph{bytes}, which means the contracted dimension must be a
multiple of $128$ for fp8 (one byte per element), of $64$ for half, and of $32$ for float. The constraint
is enforced only through a SFINAE note in the headers and appears nowhere in the specification, so a naive
fp8 port of a working half kernel fails to compile, with a misleading ``no matching constructor'' error.

Finally, the specification's own advice deserves scrutiny. It tells developers to abandon
\texttt{simdgroup\_matrix} in favour of tensors and MPP, but offers no evidence. We find that MPP does win
at every size, yet by only $1.05$ to $1.21\times$ (\cref{fig:shootout}); the deprecated path stays within
about $20\%$ and ties at small problems. The guidance is therefore directionally correct but overstated.

\begin{figure}[t]\centering
\includegraphics[width=0.74\linewidth]{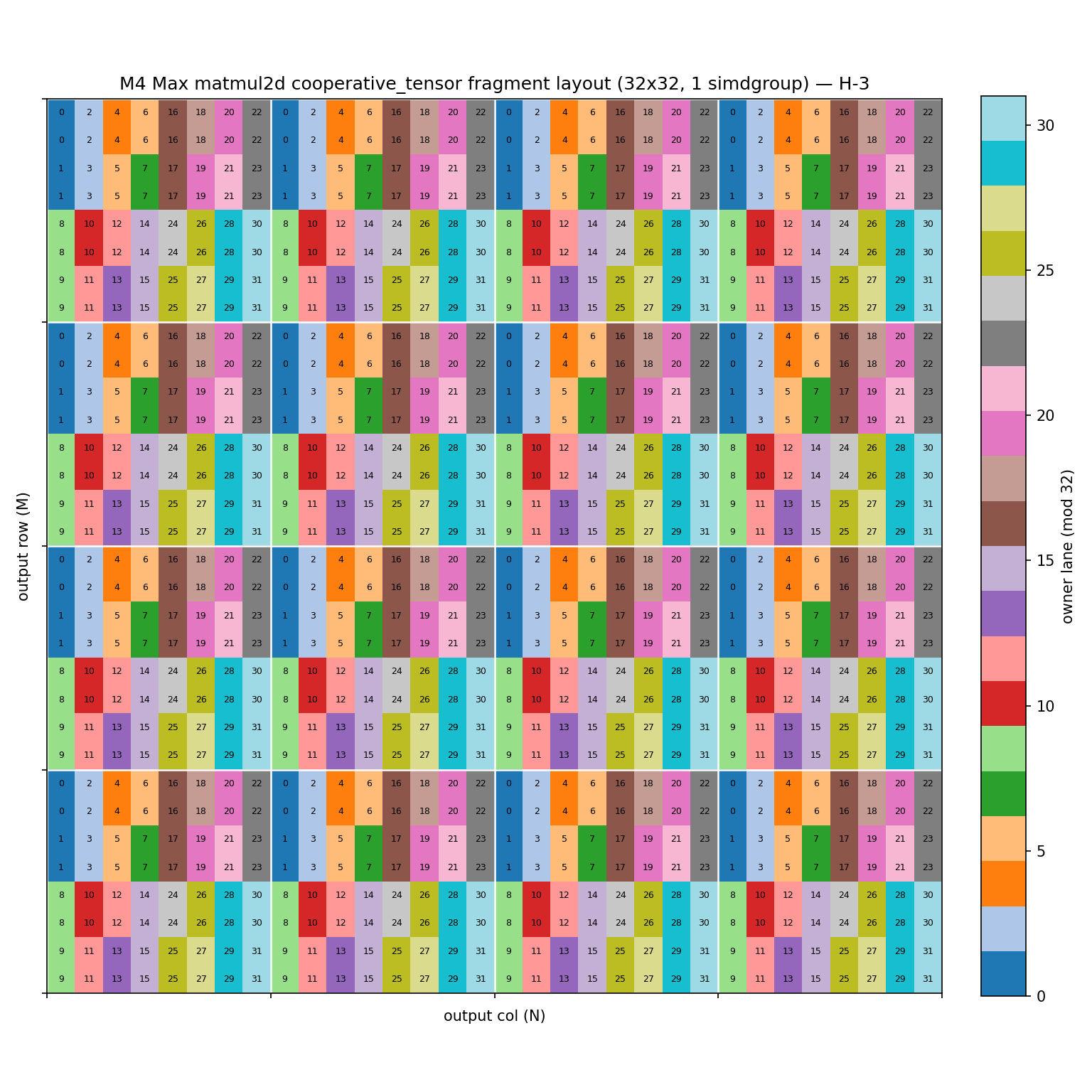}
\caption{\textbf{The reconstructed fragment layout.} The opaque \texttt{cooperative\_tensor}
partition is an $8{\times}8$ base fragment tiled across the output; each lane owns the same cell in every
$8{\times}8$ block. Inputs and output share this swizzle.}
\label{fig:frag}\end{figure}

\section{The opaque fragment layout}\label{sec:h3}

The specification declares the \texttt{cooperative\_tensor} partition ``device specific'' and documents
no layout, unlike NVIDIA's, whose WMMA fragment layout has been recovered by
microbenchmarking~\cite{suntensorcores}. We recover Apple's by a pure layout query: instantiating the
destination fragment and dumping \texttt{get\_capacity}, \texttt{is\_valid\_element}, and
\texttt{get\_multidimensional\_index} per lane, with no GEMM. For a single SIMD-group the partition is
an exact tiling: per-lane capacity $2\,(M/8)(N/8)$, the layout a repeating $8{\times}8$ base fragment
(\cref{fig:frag}) in which a lane owns two vertically adjacent rows of one column of its quadrant,
replicated across every $8{\times}8$ tile. The input (A and B) fragments share the identical
$8{\times}8$ swizzle: the partition is \emph{unified} across operands and output. This is the
Apple-Silicon analog of a documented tensor-core fragment layout, and it is the lever for layout-aware
optimization.

\section{Acting on the characterization}\label{sec:opt}

The characterization is not an end in itself; it tells us where a hand-written kernel can beat the stock
path and where it cannot. This section turns three of the findings into concrete kernel decisions: the
contiguity limits of \texttt{tensor\_inline}, the cooperative-tensor destination, and the recovered
fragment layout. It reports one optimization that clears our win bar and one that, instructively, does
not.

The first decision concerns layout. \texttt{tensor\_inline} is contiguous-only and cannot express a
strided GEMM sub-tile: its \texttt{slice()} does not offset the operation's reads, so only the $(0,0)$
tile computes correctly. A full tiled GEMM is nevertheless reachable without the host-bound MTLTensor and
MTL4 stack, by choosing a contiguous-tile $nt$ layout in which $B$ is pre-transposed and $C$ is written
tile-contiguously; this is how we obtain the $14.8$\TF baseline of \cref{fig:throughput}.

\begin{figure}[t]\centering
\includegraphics[width=0.74\linewidth]{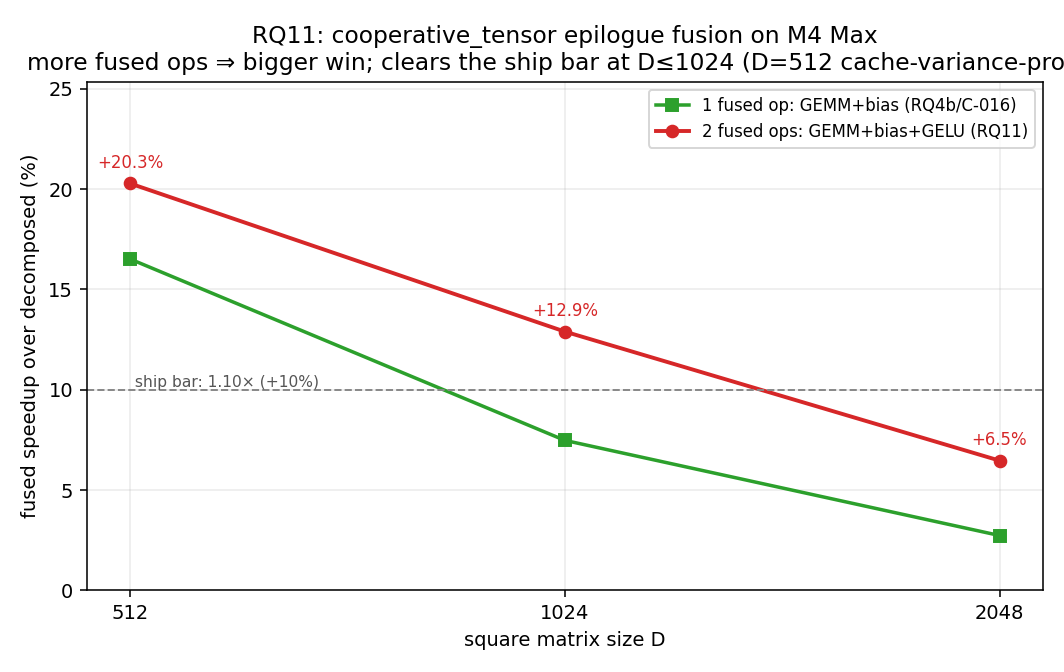}
\caption{\textbf{Epilogue fusion.} Fusing bias$+$GELU via a \texttt{cooperative\_tensor}
destination beats the decomposed path; two fused ops beat one and clear the $1.10\times$ ship bar at
$D\leq1024$.}
\label{fig:fusion}\end{figure}

The optimization that wins is epilogue fusion. Routing a \emph{bare} GEMM through a
\texttt{cooperative\_tensor} destination matches the host-bound path exactly, because there is no
per-contraction round trip to skip. The payoff is instead in the epilogue: fusing the bias and GELU
in-register on the fragment, using the recovered fragment layout, and storing the result once beats the
decomposed GEMM-then-bias-then-GELU path by $+12.9\%$ at $1024^3$ and $+6.5\%$ at $2048^3$
(\cref{fig:fusion}), which clears our $1.10\times$ bar in the size range that matters for LLM inference;
two fused operations beat one at every size. Each epilogue round trip avoided is $O(n^2)$ of memory
traffic, so the win is largest for cache-resident problems and shrinks as $O(n^3)$ compute comes to
dominate. This is the same IO-avoidance principle behind tiled-kernel compilers~\cite{triton} and
FlashAttention~\cite{flashattention}.

\begin{figure}[t]\centering
\includegraphics[width=0.74\linewidth]{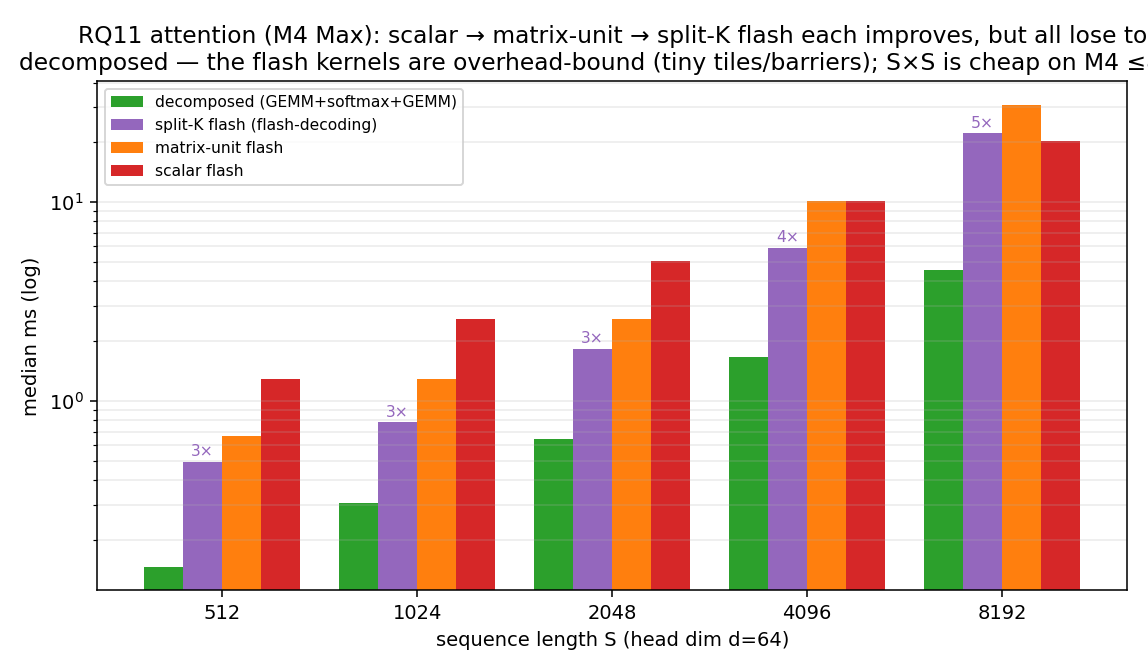}
\caption{\textbf{Attention: a negative result.} Scalar, matrix-unit, and split-K flash each improve, but
all lose to a decomposed path on M4 at $S\leq8192$: the $S{\times}S$ round trip is cheap on unified
memory and our flash kernels are overhead-bound.}
\label{fig:attn}\end{figure}

The same principle does not extend to attention, and the reason is instructive. A correct scalar
FlashAttention that uses online softmax and never materializes the $S{\times}S$ score matrix turns out to
be about $9\times$ \emph{slower} than a decomposed path that keeps the \texttt{simdgroup\_matrix} GEMMs
(\cref{fig:attn}), because fusing the softmax forces the matmuls off the matrix units and onto scalar
code. The matrix-unit and split-K FlashAttention variants we then built are correct and progressively
faster, but all remain $3.6$ to $5\times$ slower than the decomposed path: at these sequence lengths the
M4's ample unified-memory bandwidth makes the $S{\times}S$ round trip cheap, and our flash kernels are
overhead-bound by small tiles and barriers. The lesson sharpens the paper's thesis. Cooperative-tensor
fusion pays off for cheap epilogues that keep the matrix units busy, not for fusions that replace them,
and a production-grade FlashAttention for Apple Silicon~\cite{flashattention2,flashattention3} remains
open work.

\section{Related work}\label{sec:related}

\para{GPU microarchitecture reverse-engineering:} \system extends a microbenchmarking lineage that
recovers undocumented GPU internals: Jia~\etal dissect NVIDIA Volta~\cite{jiavolta} and
Turing~\cite{jiaturing} via microbenchmarks and ISA disassembly. We apply the same empirical stance to
Apple's Metal tensor path, where the obstacle is not an undocumented ISA but a deliberately opaque
high-level primitive.

\para{Tensor-core characterization:} The closest prior art characterizes NVIDIA tensor cores:
Markidis~\etal~\cite{markidis} quantify WMMA programmability, performance, and mixed-precision loss;
Sun~\etal~\cite{suntensorcores} dissect tensor-core latency, throughput, and---most relevant to us---the
numeric behavior and per-thread fragment layout of the MMA. Our results are the Apple-Silicon
counterpoint: where these works characterize a dedicated matrix unit, we find M4 Max has none
(\cref{sec:h1}), recover the analogous \texttt{cooperative\_tensor} fragment layout Apple documents
nowhere (\cref{sec:h3}), and show the fp8 row is emulated rather than accelerated (\cref{sec:h4}).

\para{Apple Silicon characterization:} Recent work benchmarks Apple Silicon at the workload level:
Feng~\etal~\cite{fengapple} profile end-to-end ML training and attribute the NVIDIA gap to system
factors; H\"ubner~\etal~\cite{hubnerapple} evaluate M1--M4 for HPC with STREAM and SGEMM FLOPS. These
measure delivered performance; none opens the Metal~4.1 tensor primitive itself---its dispatch target,
low-precision acceleration, accumulator width, or fragment layout---which is \system's contribution.

\para{Low-precision formats and kernel generation:} M4 implements the OCP FP8~\cite{ocpfp8} and
Microscaling~\cite{ocpmx} representable sets exactly; our contribution is to quantify whether the
hardware \emph{accelerates} them (it does not, on M4). Tiled-kernel compilers~\cite{triton} and the
FlashAttention line~\cite{flashattention,flashattention2,flashattention3} motivate our optimization study;
FA-3 in particular exploits Hopper fp8 tensor cores, precisely the hardware support M4 lacks.

\section{Discussion and limitations}\label{sec:disc}

Results are on a \emph{single} M4 Max under a \emph{beta} OS; cross-generation claims are out of scope
and are the natural M5 follow-up. The macOS~27.0 beta blocks direct ANE attribution (CPU/ANE power
unreported by both available tools; the ANE compiler rejects test models), so \cref{sec:h1}'s
ANE-exclusion is indirect pending a stable OS. The attention kernels are research-grade; a production
FlashAttention (register tiling, multiple SIMD-groups per block) may yet win, though M4's strong
decomposed path and ample bandwidth set a high bar.

\section{Conclusion}\label{sec:conc}

\system reverse-engineers what Metal~4.1 hides on M4 Max: \texttt{matmul2d} runs on the GPU shader path
with no dedicated matrix unit; fp8 is emulated ($0.94\times$ fp16), a footprint feature not a perf
feature; the accumulator is $\geq$fp32; the opaque $8{\times}8$ fragment layout is recovered; and
epilogue fusion built on that layout wins by $+6.5$--$12.9\%$ in the cache-resident regime. The vendor
documents an interface; we document the hardware, reproducibly, from committed code. To that end, the
microbenchmark harness, the Metal kernels, the per-cell CSVs, and the figure-regeneration scripts are all
released under the MIT license, and every figure in this paper redraws from a committed CSV through a
single \texttt{make reproduce} target.

{\small
\bibliographystyle{plainnat}
\bibliography{references}}

\end{document}